\documentclass{article}

\usepackage{microtype}
\usepackage{graphicx}
\usepackage{subfigure}
\usepackage{booktabs} 

\usepackage{hyperref}

\usepackage{amsmath}
\usepackage{amsthm}
\theoremstyle{definition}
\newtheorem{definition}{Definition}[section]
\usepackage{mathtools}
\usepackage{bbm}
\usepackage{graphicx}
\DeclareMathOperator*{\argmax}{arg\,max}
\DeclareMathOperator*{\argmin}{arg\,min}

\usepackage[accepted]{icml2021}

\icmltitlerunning{Bayesian Neural Networks with Soft Evidence}

\begin{document}

\twocolumn[
\icmltitle{Bayesian Neural Networks with Soft Evidence}



\icmlsetsymbol{equal}{*}

\begin{icmlauthorlist}
\icmlauthor{Edward Yu}{g}
\end{icmlauthorlist}

\icmlaffiliation{g}{Genesis Global Trading, New York, NY, United States}

\icmlcorrespondingauthor{Edward Yu}{eyu@genesistrading.com}

\icmlkeywords{Machine Learning, ICML}

\vskip 0.3in
]



\printAffiliationsAndNotice{} 

\begin{abstract}
Bayes's rule deals with hard evidence, that is, we can calculate the probability of event $A$ occuring given that event $B$ has occurred. Soft evidence, on the other hand, involves a degree of uncertainty about whether event $B$ has actually occurred or not. Jeffrey's rule of conditioning provides a way to update beliefs in the case of soft evidence. We provide a framework to learn a probability distribution on the weights of a neural network trained using soft evidence by way of two simple algorithms for approximating Jeffrey conditionalization. We propose an experimental protocol for benchmarking these algorithms on empirical datasets and find that Jeffrey based methods are competitive or better in terms of accuracy yet show improvements in calibration metrics upwards of 20\% in some cases, even when the data contains mislabeled points.
\end{abstract}

\section{Introduction}
In a traditional probabilistic reasoning framework, we make inferences based on \emph{hard evidence}, or in other words, observation of some event $B$. Then we may be able to say something about the likelihood of another event $A$ by calculating $P(A|B)$. Take the classic case of Bernoulli trials, in which we are flipping a (possibly biased) coin multiple times. Each flip of the coin constitutes hard evidence, as we are certain whether the coin landed heads or tails, and we can record the event definitively: either $B=1$ or $B=0$. Then we may use a traditional Bayesian framework to estimate the probability the next flip will result in heads or tails. 

In contrast, \emph{soft evidence} arrives in the form of another probability distribution $R(B)$ as described by \citet{peng}, or more simply, as an ``unreliable testimony" of $B$ accompanied by a strength of belief in the $[0, 1]$ interval as in \citet{chan2005revision} and \citet{jacobs}. In the example of Bernoulli trials, imagine that in poor visibility conditions we are not able to ascertain definitively whether the coin landed heads or tails, but we estimate that there is a 80\% probability it landed heads, i.e., $R(B = 1) = .8$. Soft evidence appears in many real world contexts, from unreliable witness testimonies in murder trials \citep{fields2013toward} to game theoretic decision making \citep{dietrich2016belief}. One problem setting that we focus on in this paper is that of crowdsourced data, where each data point has multiple annotators, yet there is no ``true" label available, only a categorical distribution over possible labels \citep{rodrigues2013learning}. Logically it seems there should be a way to make use of soft evidence that takes into account the uncertainty of the label, in fact, it seems almost wasteful and anti-Bayesian to throw away the uncertainty and train with a point estimate (hard label).

However, since Bayes' theorem cannot directly be applied, there is no consensus on how to update one's beliefs in order to calculate $P(A|B)$, or in a generalized machine learning setting, how to train models given soft evidence. One sensible approach is Jeffrey's rule of conditioning \citep{jeffrey1990logic} \citep{shafer1981jeffrey}, which can be thought of as an modification of the posterior probability distribution given a set of constraints. For a Bernoulli constraint, Jeffrey's rule reduces to standard probabilistic reasoning in the case where we observe $B$ with probability $R(B) = 1$, and is otherwise a convex combination of $P(A|B=1)$ and $P(A|B=0)$. Although it seems natural to interpret soft evidence in this way, it is of philosophical debate (beyond the scope of this paper) whether it is ``correct". Suffice it to say that there are both many extensions of Jeffrey's original rule \citep{ichihashi1989jeffrey} \citep{benferhat2010framework} \citep{benferhat2011jeffrey} \citep{smets1993jeffrey}, and alternative approaches based on the broader field of Dempster-Shafer theory \citep{shafer1992dempster} \citep{denoeux2019logistic}.

There are several related areas of study when it comes to applying soft evidence to practical machine learning domains. Pearl's method is another method of dealing with soft evidence, whereby a Bayesian network is extended with an additional node, and soft evidence is recast as hard evidence of a virtual event. In fact, \citet{chan2005revision} show that one may convert between Pearl's method and Jeffrey's method by specifying likelihood ratios. \citet{peng} extend this work from theory to application by providing several algorithms for approximating Jeffrey's method on Bayesian networks. We in turn build upon \citet{peng} and extend the algorithms for usage in neural networks. 

A related field is that of learning noisy labels, where data is given in the form $\{x_i, y_i \sim R(\gamma)\}$, however unlike the soft evidence case where $R(\gamma_i)$ is given, in the noisy label case only $y_i$ is observed. \citet{rolnick2017deep} posit that under certain types of noise, the noisiness of the labels can be ignored and a neural network is trained as if the noise did not exist. \citet{karimi2020deep} give a survey of methods for dealing with label noise, from label preprocessing to weighted loss functions. Since soft evidence can be converted to a noisy label by taking $y_i = \argmax_{\gamma} R(\gamma)$, we use methods from learning noisy labels as baselines.

The main contributions of this paper are:
\begin{enumerate}
    \item To our knowledge, we are the first to extend Jeffrey conditionalization to the training of neural networks. We provide end-to-end algorithms, including two approximation methods for Jeffrey's method: one simple ensemble method and one method based upon the \emph{Bayes by Backprop} method \citep{blundell2015weight}. 
    \item We benchmark against methods in related domains, including baselines such as ensembles which ignore the soft evidence and training methods for noisy labels. We show empirical success in reducing calibration error, while maintaining equal or better levels of accuracy.
\end{enumerate}

The rest of the paper is organized as follows: Section \ref{sec:jeffrey} defines Jeffrey's rule and gives an example. Section \ref{sec:algo_overview} gives two approximation algorithms for learning neural networks using Jeffrey's rule. Section \ref{sec:experiments} defines the experimental protocol, including evaluation criteria, methodology for generating corrupted data, baseline models, and results. Section \ref{sec:conclusion} concludes.

\section{Jeffrey's Rule of Conditioning} \label{sec:jeffrey}
In this section we explore the preliminaries behind Jeffrey's rule. The following two definitions are due to \citet{chan2005revision}. Consider a set of mutually exclusive and exhaustive events $\gamma_1, ..., \gamma_n$. We wish to update a probability distribution $P(\alpha)$, given a set of soft evidence constraints of the form $R(\gamma_i) = q_i \text{ for } i = 1, 2, ...,n$.

\theoremstyle{definition}
\begin{definition}[probability kinematics]
Suppose there are two distinct probability distributions $P$ and $P'$, that is, it is the case that $P(\gamma_i) \neq P'(\gamma_i)$ for at least one $i$. $P'$ is obtained from $P$ by probability kinematics on $\gamma_1, ..., \gamma_n$ iff for every event $\alpha$ in the probability space
$$P(\alpha | \gamma_i) = P'(\alpha | \gamma_i) \text{ for } i = 1, 2, ...,n$$
\end{definition}

\begin{definition}[Jeffrey's Rule]
Jeffrey's rule produces the sole distribution $J$ that is obtained from $P$ by probability kinematics on $\gamma_1, ..., \gamma_n$:
\begin{equation}
\begin{split}
    J(\alpha) &= \sum_{i=1}^n R(\gamma_i) \frac{P(\alpha, \gamma_i)}{P(\gamma_i)} \\
    &= \sum_{i=1}^n P(\alpha | \gamma_i) R(\gamma_i)
\end{split}
\end{equation}
\theoremstyle{definition}
\end{definition}

Rephrased, Jeffrey's rule finds the closest distribution (minimum KL-divergence \citep{peng}) to $P$ that satisfies the constraints $R$. If there is no uncertainty, i.e., $R(\gamma_1) = 1$, then $J(\alpha) = P(\alpha)$, and Jeffrey conditionalization does not update $P(\alpha)$. Otherwise, the posterior is a mixture of all possibilities of $\gamma$. 
\section{Jeffrey's and Neural Networks} \label{sec:algo_overview}
Now consider a Bayesian neural network with weights $w$ and data $D_k = \{(x_i, y_i \sim R(\gamma))\}$. The posterior predictive distribution is 
\begin{equation}
    p(y|x, D_k) = \int p(y|x,w)p(w|D_k)dw
\end{equation}
This neural network is dependent on one particular instantiation of the data. In a standard setting with fixed $(x, y)$, this is the only way to proceed. In contrast, a Jeffrey neural network would depend on all possible instantiations of the data:

\begin{equation}
    J(y|x) = \sum_k p(y|x, D_k) R(D_k)
\end{equation}

Unfortunately this is computationally intractable as this would require enumerating through all instantiations of $D_k$, and training a separate neural network for each possibility of $D_k$. We give the following two algorithms as practical alternatives.

\subsection{Sparse K Approximation}
We limit ourselves to training $K$ neural networks as computational resources permit.
\begin{algorithm}
 \caption{Sparse K Approximation}
 \label{algo1}
\begin{algorithmic}
   \FOR{$k=1$ {\bfseries to} $K$}
   \STATE Draw sample $D_k \sim R(D)$
   \STATE Train Bayesian neural network $p(y|x, D_k)$
   \ENDFOR
   \STATE Ensemble is then $\bar{J}(y|x) = \sum_k p(y|x, D_k)$
\end{algorithmic}
\end{algorithm}
This is a Monte Carlo direct sampling method. It is apparent that $\lim_{k \to \infty} \bar{J}(y|x) = J(y|x)$, as every possible instantiation of $D_k$ will be sampled a proportional number of times according to $R(D_k)$. In the case where $k$ is finite, this can be viewed as a generalization of \emph{bagging} \citep{breiman1996bagging}: we create bootstrap samples of data to train on, and average the predictors. Whereas in the bagging procedure we sample uniformly from the empirical distribution $\widehat R(D)$, here we sample from the true distribution $R(D)$.

\subsection{Jeffrey's By Backprop} \label{sec:jnn}
\citet{blundell2015weight} introduced the now widely adopted \emph{Bayes by Backprop} algorithm to train Bayesian neural networks. The idea is to use variational inference to find a distribution on weights that approximate the true Bayesian posterior. Using only a minor modification to the loss function, we introduce a similar \emph{Jeffrey's by Backprop} method whose output will be a probability distribution over weights. We introduce a distribution over weights $q(w|\theta)$ parameterized by $\theta$. The objective of the \emph{Bayes by Backprop} method is to learn the optimal parameters $\theta$ that minimize the Kullback-Leibler divergence between $q(w|\theta)$ and $P(w|D)$.
\begin{equation}
    \theta^* = \argmin_\theta \text{KL}[q(w|\theta) || P(w|D)]
\end{equation}

KL divergence is defined as:
\begin{equation}
    \text{KL}[q || p] = \int q(x) \log\left(\frac{q(x)}{p(x)}\right)\ dx
\end{equation}

In the spirit of Jeffrey's rule, we modify the objective to minimize the sum of KL divergences across all possible $P(w|D_k)$.
\begin{align}
    \theta^* &= \argmin_\theta \sum_k \text{KL}[q(w|\theta) || P(w|D_k)] R(D_k) \\
    &= \argmin_\theta \left[ \sum_k \int q(w|\theta) \log \frac{q(w|\theta)}{P(w)P(D_k|w)}dw\right] R(D_k) \\
    &= \argmin_\theta \sum_k \Big( \text{KL}[q(w|\theta)||P(w)] \\ 
    &\phantom{\argmin_\theta \sum_k ( KL} - E_{q(w|\theta)}[\log P(D_k|w)] \Big) R(D_k) \\
    &= \argmin_\theta \text{KL}\left[q(w|\theta)||P(w)\right] \\ 
    &\phantom{\argmin_\theta} - \sum_k E_{q(w|\theta)}[\log P(D_k|w)] R(D_k)
\end{align}
This leads to the loss function
\begin{align}
    F(D, \theta) &= \text{KL}[q(w|\theta)||P(w)] \\
     &\phantom{\text{KL}} -\sum_k E_{q(w|\theta)}[\log P(D_k|w)] R(D_k)
\end{align}
and the approximate loss
\begin{align}
    F(D, \theta) \approx \sum_{i=1}^n \Big[ \log q(w^{(i)}|\theta)-\log P(w^{(i)}) \\
    \phantom{} - \log P(D^{(i)}|w^{(i)}) \Big]
\end{align}
where $w^{(i)}$ is a Monte Carlo sample of weights drawn from the variational posterior $q(w|\theta)$, and $D^{(i)}$ is sampled from $R(D)$. With the modified loss, we may proceed in the exact same manner as outlined in \citep{blundell2015weight}, and train by backpropagation. 
\section{Experiments} \label{sec:experiments}
To ensure that results are reproducible, code and data for experiments are available on our Github at \url{https://github.com/edwardyu/soft-evidence-bnn}.

\subsection{Evaluation Criteria}
Our goal is not to beat state-of-the-art benchmarks on the datasets, but rather to show an improvement in both/either accuracy or calibration metrics over the baseline models. A well calibrated model is one that is correct on high confidence predictions and incorrect on low confidence predictions. We use two \emph{proper scoring rules} to measure calibration, the negative log-likelihood score (NLL) and the Brier score \citep{guo2017calibration} \citep{lakshminarayanan2016simple}. The negative log likelihood is defined as 
\begin{equation}
    \mathcal{L} = - \log J(y | x) 
\end{equation}
And the Brier score is
\begin{equation}
    \mathcal{L} = \frac{1}{C} \sum_{c=1}^C (R(y=c) - J(y=c|x))^2
\end{equation}
All errors reported in parentheses are 1 standard deviation.

\subsection{Models and Baselines}
For all methods, we set a common base learner: a neural network with a fixed architecture trained via the \emph{Bayes by Backprop} algorithm \citep{blundell2015weight}. The specific implementation is based on an open source library created by \citet{esposito2020blitzbdl}. We compare the following methods:
\begin{enumerate}
    \item Sparse K approximation to Jeffrey's (\textbf{SparseK}): see algorithm \ref{algo1}.
    \item Jeffrey's by Backprop (\textbf{JNN}): see section \ref{sec:jnn}.
    \item Noisy labels (\textbf{NL}): train on $D = \{(x_i, \argmax_\gamma R(\gamma_i))\}$. In any hard label dataset this is the default assumption.
    \item Noisy labels ensemble (\textbf{NLE}): train on $D = \{(x_i, \argmax_\gamma R(\gamma_i))\}$ using K base learners, and output the majority vote class.
    \item Bagging (\textbf{Bag}): Sample data from empirical distribution $\widehat R(D)$, train on sampled data.
\end{enumerate}

All methods are trained for 100 epochs, and we set $K=3$ for ensemble methods.
\subsection{CIFAR-10H} \label{corrupt}
The CIFAR-10H dataset consists of 10,000 color images belonging to 10 possible classes \citep{peterson2019human}. It is a subset of the popular CIFAR-10 dataset; the difference is that CIFAR-10H is soft labeled, with each image having multiple annotations. In total the authors collected  511,400 human labels, but removed annotators who scored low on attention-checks. On average, the final dataset is not very noisy: the mean vote share for the most popular label is $95.44\%$. We train using a Bayesian version of the LeNet architecture \citep{lecun1998gradient}. 

Table \ref{table:1} shows results. The relatively low accuracies across the board highlights the difference in calibration. When comparing to hard label method \textbf{NL}, all other methods significantly reduce the NLL and Brier scores. However, it should be noted that \textbf{JNN} is a single network method, and thus uses much less compute resources than the ensemble methods. \textbf{SparseK} achieved the highest accuracy and lowest calibration error.

\begin{figure}
    \centering
    \includegraphics[width=0.25\textwidth]{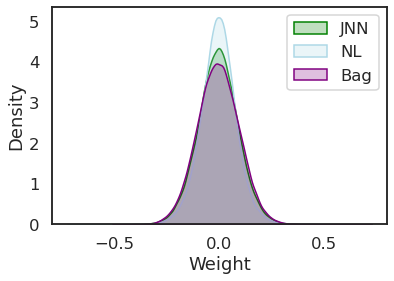}
    \caption{A histogram of weights from models trained on the corrupted CIFAR-10 dataset. JNN and Bag models have increased uncertainty in the weights due to training on soft labels. The NL method first converts the soft label to a hard label.}
    \label{fig:density}
\end{figure}

\begin{table}[h!]
\centering
\caption{Results on CIFAR-10H}
\scalebox{.9}{
\begin{tabular}{|c | r | r | r |} 
 \hline
 Model & Accuracy & NLL $\times 10$ & Brier $\times 10^3$ \\ 
 \hline
 SparseK & $\mathbf{47.47 (\pm 0.68)}$ & $\mathbf{1.91 (\pm 0.02)}$ & $\mathbf{7.47 (\pm 0.08)}$\\ 
 JNN & $46.98 (\pm 0.37)$ & $2.63 (\pm 0.29)$ & $8.78 (\pm 0.43)$\\
 \hline
 NL & $47.02 (\pm 0.65)$ & $3.93 (\pm 0.14)$ & $9.85 (\pm 0.03)$\\
 NLE & $45.93 (\pm 1.83)$ & $2.03 (\pm 0.06)$ & $7.76 (\pm 0.25)$\\
 Bag & $46.13 (\pm 0.47)$ & $1.97 (\pm 0.04)$ & $7.64 (\pm 0.14)$\\
 \hline
\end{tabular}
}
\label{table:1}
\end{table}

\subsection{LabelMe}
The LabelMe dataset is an image classification dataset that appears in many forms, here we use a subset obtained from \citet{rodrigues2018deep}. The dataset consists of 2688 images, each corresponding to one of the following categories: highway, inside city, tall building, street, forest, coast, mountain or open country. This is a soft evidence dataset: each image was labeled by multiple annotators on the Amazon Mechanical Turk platform. On average each image has $2.547$ labels and the annotators have a mean accuracy of $69.2\%$. When training on this dataset, we used the very deep convolutional network architecture (VGG-11) as proposed by \citet{simonyan2014very}.

Table \ref{table:2} shows results. The labels in this dataset are noisier than CIFAR-10H, and Jeffrey-based methods show an increase in accuracy compared to their Bayesian counterparts. As before, any ensemble method or \textbf{JNN} reduces the calibration error when compared with the default hard label method \textbf{NL}.

\begin{table}[h!]
\centering
\caption{Results on LabelMe}
\scalebox{.9}{
\begin{tabular}{|c | r | r | r |} 
 \hline
 Model & Accuracy & NLL $\times 10$ & Brier $\times 10^3$ \\ 
 \hline
 SparseK & $70.73 (\pm 1.60)$ & $1.35 (\pm 0.03)$ & $8.16 (\pm 0.18)$\\ 
 JNN & $\mathbf{70.76 (\pm 0.97)}$ & $\mathbf{1.19 (\pm 0.05)}$ & $\mathbf{6.70 (\pm 0.21)}$\\
 \hline
 NL & $64.11 (\pm 1.94)$ & $2.22 (\pm 0.32)$ & $8.87 (\pm 0.29)$\\
 NLE & $66.89 (\pm 0.56)$ & $1.33 (\pm 0.06)$ & $7.82 (\pm 0.39)$\\
 Bag & $67.99 (\pm 2.52)$ & $1.37 (\pm 0.07)$ & $8.03 (\pm 0.37)$\\
 \hline
\end{tabular}
}
\label{table:2}
\end{table}

\section{Conclusions and Future Work} \label{sec:conclusion}
In this paper we showed two methods for training Bayesian neural networks when the dataset contains soft evidence, where labels are uncertain and are only presented as probabilistic inputs. In contrast to previous approaches such as converting to hard labels or bagging, Jeffrey conditionalization is provably optimal in the sense that it follows the principle of probability kinematics. Evaluating on several real-world datasets, we show that Jeffrey based methods are competitive in terms of accuracy, while being better calibrated and showing lower KL-loss. Since our examples involve human labelers, the provided empirical $R(\gamma)$ are likely to differ from the true data distribution. Future work may focus on understanding this scenario better: are there bounds we can place on the loss of our model if $\text{KL}[R(\gamma) || R^{\text{true}}(\gamma)]$ is bounded? Additionally, it may be further worth exploring the relationship between Jeffrey conditionalization and bagging. Does the ample theoretical work on the Jeffrey posterior provide clues as to why bagging works so well in practice?
\bibliography{citations}
\bibliographystyle{icml2021}

%
%
%

\end{document}